\documentclass[10pt,twocolumn,letterpaper]{article}

\usepackage[pagenumbers]{cvpr} 

\usepackage{graphicx}
\usepackage{amsmath}
\usepackage{amssymb}
\usepackage{booktabs}
\usepackage{caption}
\usepackage{subcaption}
\usepackage{mathtools}

\DeclarePairedDelimiter\floor{\lfloor}{\rfloor}
\usepackage[pagebackref,breaklinks,colorlinks]{hyperref}

\usepackage[capitalize]{cleveref}
\crefname{section}{Sec.}{Secs.}
\Crefname{section}{Section}{Sections}
\Crefname{table}{Table}{Tables}
\crefname{table}{Tab.}{Tabs.}

\def\cvprPaperID{*****} 
\def\confName{CVPR}
\def\confYear{2022}

\begin{document}

\title{NAAP-440 Dataset and Baseline for Neural Architecture Accuracy Prediction}

\author{Tal Hakim \\
Smart Shooter, Kibbutz Yagur, Israel \\
{\tt\small tal.hakim@smart-shooter.com}
}
\maketitle

\begin{abstract}
Neural architecture search (NAS) has become a common approach to developing and discovering new neural architectures for different target platforms and purposes. However, scanning the search space is comprised of long training processes of many candidate architectures, which is costly in terms of computational resources and time. Regression algorithms are a common tool to predicting a candidate architecture's accuracy, which can dramatically accelerate the search procedure. We aim at proposing a new baseline that will support the development of regression algorithms that can predict an architecture's accuracy just from its scheme, or by only training it for a minimal number of epochs. Therefore, we introduce the NAAP-440 dataset of 440 neural architectures, which were trained on CIFAR10 using a fixed recipe. Our experiments indicate that by using off-the-shelf regression algorithms and running up to 10\% of the training process, not only is it possible to predict an architecture's accuracy rather precisely, but that the values predicted for the architectures also maintain their accuracy order with a minimal number of monotonicity violations. This approach may serve as a powerful tool for accelerating NAS-based studies and thus dramatically increase their efficiency. The dataset~\footnote{\url{https://www.kaggle.com/datasets/talcs1/naap-440}} and code used in the study~\footnote{\url{https://github.com/talcs/NAAP-440}} have been made public. 


\end{abstract}

\begin{figure}[]
\centering
  \includegraphics[width=\linewidth]{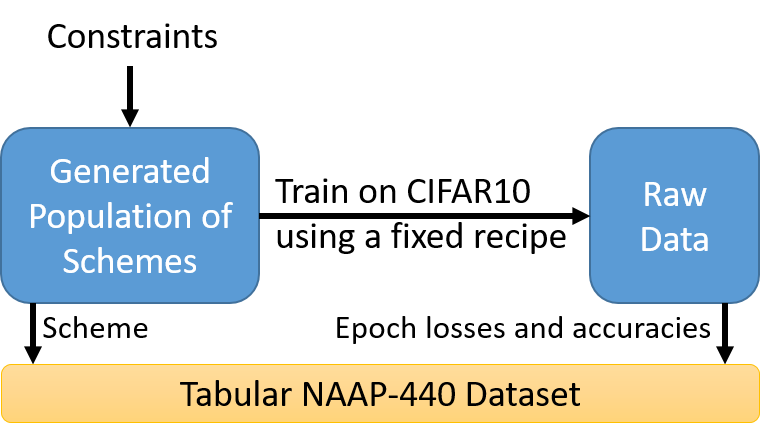}
  \caption[]{The creation of the dataset starts with defining a population of neural architectures. All the architectures are trained using the same fixed recipe and are evaluated on CIFAR10 test set after every epoch. The NAAP-440 dataset is formed from the collected data.}
  \label{fig:approach}
\end{figure}

\section{Introduction}
\label{sec:intro}
Over the recent decade, convolutional neural networks have been evolving and improving in their accuracy and efficiency~\cite{krizhevsky2017imagenet,simonyan2014very,he2016deep,szegedy2016rethinking,xie2017aggregated,sandler2018mobilenetv2,huang2017densely, hu2018squeeze}, with the development of new ideas and approaches, such as batch normalization~\cite{ioffe2015batch}, residual blocks~\cite{he2016deep}, inverted residual blocks~\cite{sandler2018mobilenetv2}, grouped convolutions~\cite{xie2017aggregated}, squeeze-and-excitation blocks~\cite{hu2018squeeze} and many others. In the recent years, a leading approach for discovering new architectures is Neural Architecture Search (NAS)~\cite{zoph2016neural,zoph2018learning,tan2019mnasnet,howard2019searching,tan2019efficientnet,tan2021efficientnetv2} and other NAS-inspired approaches~\cite{radosavovic2020designing,liu2022convnet}, in the sense that they are based on a massive experimentation with many candidate architectures.

The main drawback of the NAS approach is its resource consumption, which for instance, could be 500 GPUs running for 4 days~\cite{zoph2018learning}. While it is a rather trivial and inexpensive task to evaluate a candidate architecture in terms of efficiency, by counting MACs or measuring its runtime on a single batch~\cite{tan2019mnasnet,tan2019efficientnet,tan2021efficientnetv2}, it is a rather expensive task to compute the architecture's accuracy, as it requires training it first using a given recipe.

The idea of predicting an architecture's accuracy has already been studied~\cite{zhou2020performance,istrate2019tapas,baker2017practical,long2020performance,mellor2021neural}. The NAS-Bench-101~\cite{ying2019bench}, NAS-Bench-201~\cite{dong2020bench} and NAS-Bench-301~\cite{siems2020bench} datasets of neural architectures and their corresponding accuracies have been published over the years, which have been extensively used for designing regression algorithms that are intended for accelerating the search process, as well as evaluating search techniques~\cite{wen2020neural,white2021powerful,luo2020accuracy,xu2021renas,gracheva2021trainless,zela2020bench}.

We aim at introducing a new baseline for developing regression algorithms and a set of features that will allow for correctly predicting a candidate architecture's accuracy with a minimal cost of resources. By saving 90\% up to 100\% of an architecture's training time, we anticipate that our approach will join the efforts to make NAS dramatically more efficient, which will allow for the scan of wider search spaces and thus the discovery of better architectures. 

For that purpose, we introduce the NAAP-440 dataset, which consists of 440 neural architectures trained and evaluated on the CIFAR10 dataset~\cite{krizhevsky2009learning}. As visualized in Figure~\ref{fig:approach}, we generate 440 neural architectures using a set of constraints and use a fixed training recipe to train each of them. After each epoch, we document the value of the loss function and evaluate the model on the CIFAR10 test set. As a result, our dataset contains both scheme features and features from the training process. We evenly sample 40 architectures, to form a training set of 400 architectures and a test set of 40 architectures. Our aim is that, given a minimal number of features from the training epochs, to predict an architecture's highest accuracy score achieved on the CIFAR10 test set, over all its training epochs. We perform ablation studies on the feature set and provide a baseline of performance reports using many regression algorithms and combinations of features. We measure the quality of the accuracy prediction using two measures: mean absolute error and monotonicity score.

Our experiments indicate that by using off-the-shelf regression algorithms, it is actually possible not to train the architectures on CIFAR10 at all, or in other words, to accelerate the training process by 100\%, and still predict their CIFAR10 test set accuracy scores with MAE of 0.5\% and only 42 monotonicity violations, out of $\binom{40}{2} = 780$ possible violations. Furthermore, the experiments indicate that by only training each architecture on CIFAR10 for 10\% of the epochs and evaluating them after every epoch, or in other words, accelerating the training process by 90\%, it is possible to predict their CIFAR10 test set accuracy scores with MAE of 0.4\% and only 25 monotonicity violations, as demonstrated in Figures~\ref{fig:predictionResults},\ref{fig:monotonicityResults} and Table~\ref{tbl:baseline}. We believe that using this approach, it will be possible to accelerate any NAS-based solution, by building similar regression algorithms using a rather small population of architectures sampled from the search space, then using the regression algorithms for the whole search space. 

Our contributions in this study are as follows. First, we create the NAAP-440 dataset and make it public for further research. Second, we conduct ablation studies and experiments that form a baseline of performances to be used as a reference. Third, we introduce the monotonicity-based evaluation and show that evaluating candidate architectures can be accelerated by up to 100\% and still be accurate.

We encourage researchers to use the NAAP-440 dataset in their studies and try to improve it from any aspect, including improving the reported baseline results, extending the evaluation metrics and extending the dataset.

\begin{figure*}[]
\centering
  \includegraphics[width=\linewidth]{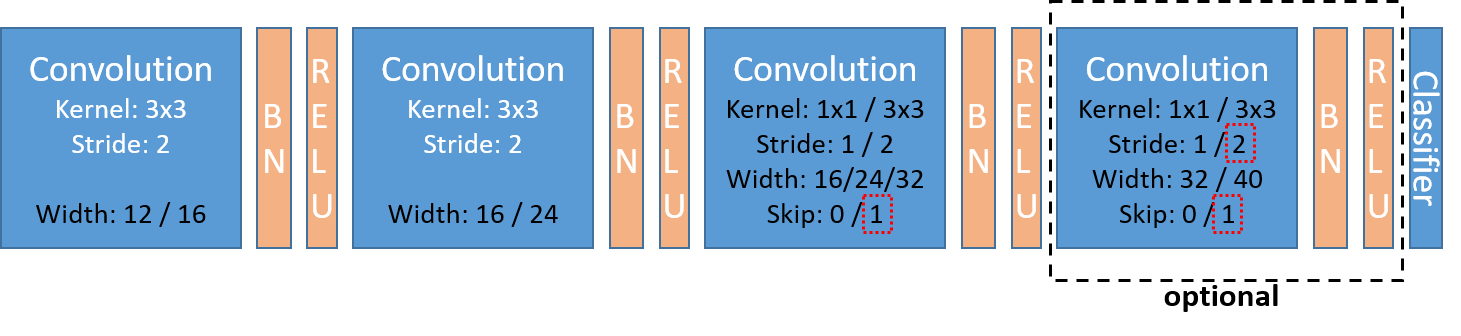}
  \caption[]{The set of constraints that define the population of network schemes that form the dataset. A red dotted rectangle denotes an assignment that is only possible under certain conditions: skip is only possible if a layer's input and output shapes match, stride=2 at the forth layer is only possible if the third layer has been assigned with stride=1, as the number of stages has been defined to be 2 or 3.}
  \label{fig:scheme_generation}
\end{figure*}

\section{Dataset}
\label{sec:dataset}

\paragraph{Scheme Generation.}
The schemes are generated using a DFS scan on the convolutional layers with a set of constraints. As visualized in Figure~\ref{fig:scheme_generation}, some constraints are imposed on the whole model, while others are imposed per layer. For the model itself, the number of convolutional layers varies between 3 and 4 and the number of stages varies between 2 and 3. The layers themselves can vary in their kernel size, width, stride and whether they have a skip connection. Every convolutional layer is followed by batch normalization~\cite{ioffe2015batch} and ReLU. We always use convolutional layers with no bias and with padding of $\floor{\frac{S}{2}}$, where $S$ is the kernel size. Since layers 1,2 are fixed to have stride=2, only up to one of layers 3,4 can be assigned with stride=2, to enforce the number of stages varying between 2 and 3. A skip connection can be applied only if the layer's input shape agrees with its output shape, which can only occur when the stride is selected to be 1 and the width of the current layer is equal to the width of the previous layer. The total number of valid schemes according to our set of constraints is 440.

\paragraph{Architecture Training.}
In order to generate the dataset, we turn the 440 schemes into actual architectures and train each of them on the CIFAR10 dataset for 90 epochs. For reproducibility, we initialize a fixed random-seed before initializing each model's weights and before initializing the shuffled training set loader, as well as switching the accelerators to deterministic mode. This ensures that all the architectures are trained using the same permutation of the training set. We use CE loss and the SGD optimizer with warm restarts~\cite{loshchilov2016sgdr} every 3 epochs. The training recipe is as follows: batch-size: 256, momentum: 0.9, weight-decay: 0.0001, learning-rate: 0.1, with exponential decay, by multiplying by 0.1 after every epoch. As the whole training runs for 90 epochs and the learning rate restarts after every 3 epochs, the process can be described as a sequence of 30 training cycles, each consisting of 3 epochs.

\paragraph{Dataset Structure.}
We build a tabular dataset that contains all the 440 architectures. Each architecture is reported with its scheme details and the collected data from its training and evaluation on CIFAR10. The scheme details include the network's depth, number of stages, first and last layer widths, number of total parameters and number of total MACs. The data from the training and evaluation on CIFAR10 is provided for each of the 90 epochs and includes the model's accuracy on CIFAR10 test set after each epoch is completed, as well as each epoch's mean and median training loss over all the SGD batches.

\begin{figure}[]
\centering
  \includegraphics[width=\linewidth]{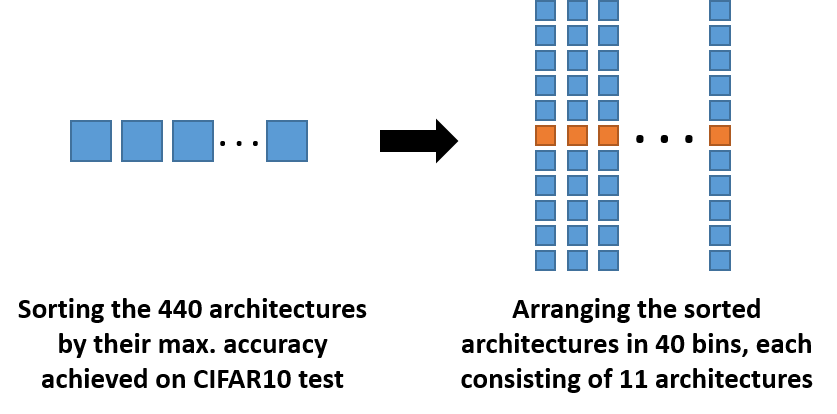}
  \caption[]{Division of the NAAP-440 dataset into train and test sets. The architectures at the central index of each bin are allocated to the test set.}
  \label{fig:trainTestDivision}
\end{figure}

\paragraph{Separating into Train and Test Sets.}
We would like to create a fair separation of the 440 architecture data records into train and test sets. Our first choice deals with quantities - we would like to allocate 400 samples to the training set and 40 samples to the test set. The second and more complicated choice is how to select the 40 test samples in a deterministic and even way, in terms of the network structure and performances. The original order of the architectures has been determined by the DFS scan over the set of constraints, which we consider unsatisfactory. Instead, as visualized in Figure~\ref{fig:trainTestDivision}, we sort the architectures by their accuracy scores on the CIFAR10 test set, using the highest achieved score over all the training epochs. Then, we arrange the 440 architectures in 40 bins, such that each bin consists of 11 architectures that have similar accuracy scores on the CIFAR10 test set. From each bin, we allocate the architecture at the central index to the test set. This deterministic operation leaves us with 400 training samples and 40 representative test samples.

\section{Evaluation Metrics}
\label{sec:evaluation_metrics}

\begin{figure}[]
\centering
  \includegraphics[width=\linewidth]{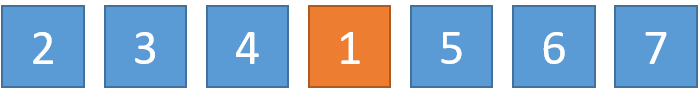}
  \caption[]{A sequence of seven items ordered with a single misplaced item, causing three violations of monotonicity. The maximum number of violations possible is the number of unordered pairs, which is equal to $\binom{7}{2} = 21$ in this example. The monotonicity score of this example will therefore be $1 - \frac{3}{21} = \frac{6}{7}$}
  \label{fig:monotonicity}
\end{figure}

\paragraph{Prediction Accuracy.}
In order to evaluate the quality of a proposed regression model, we use multiple measures. The first and rather obvious one is the Mean Absolute Error (MAE) metric, which reflects the overall quality of the regression model.

\paragraph{Monotonicity Score.}
Being able to compare two candidate architectures to each other is an important capability when searching for neural architectures. Using a regression model's predicted accuracy of the two architectures can serve as a measure for comparison. However, while some proposed regression models might provide accurate predictions for most of the test samples, their predictions are not guaranteed to preserve the original order of the samples, which may lead to a misleading comparison. For that reason, another metric that we propose is a monotonicity score. As visualized in Figure~\ref{fig:monotonicity}, given $N$ samples, the maximum possible number of monotonicity violations, is the number of pairs, which is $\binom{N}{2} = N(N-1)/2$. Therefore, the monotonicity score we propose is

\begin{equation}
1 - \frac{\#violations}{\binom{N}{2}} ,
\end{equation}

\noindent while in our case, $N=40$, such that $\binom{N}{2} = 780$. Thus, the monotonicity score will be equal to 1 when the predictions maintain the order of the architecture accuracies perfectly. In contrast, it will be equal to 0 when the predictions result in an exactly reversed version of the original order of the architecture accuracies, a situation where every pairwise comparison of architectures using the regression model will be incorrect.

\paragraph{Search Acceleration.}
The architecture accuracy prediction can be performed under different conditions. Since our aim is to predict the highest test accuracy on CIFAR10 in the course of 90 training epochs, a regression model that only uses an architecture's details from the 9 first epochs, which are 10\% of the training process, accelerates the search by 90\%. Similarly, a regression model that only uses the schematic features accelerates the search by 100\%, as it requires no training at all. We therefore evaluate different regression algorithms on the dataset under different feature sets, to emphasize the trade-off between search acceleration, prediction accuracy and prediction monotonicity.

\begin{figure}
     \centering
     \begin{subfigure}[b]{0.4\textwidth}
         \centering
         \includegraphics[width=\textwidth]{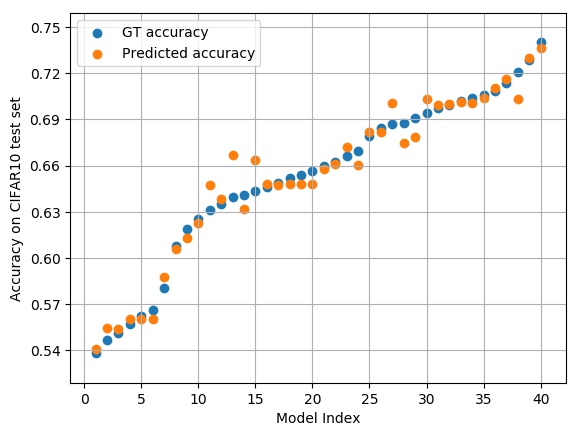}
         \caption{Using Gradient Boosting (N=100 estimators) and only scheme features as input, MAE=0.006.}
     \end{subfigure}
     \begin{subfigure}[b]{0.4\textwidth}
         \centering
         \includegraphics[width=\textwidth]{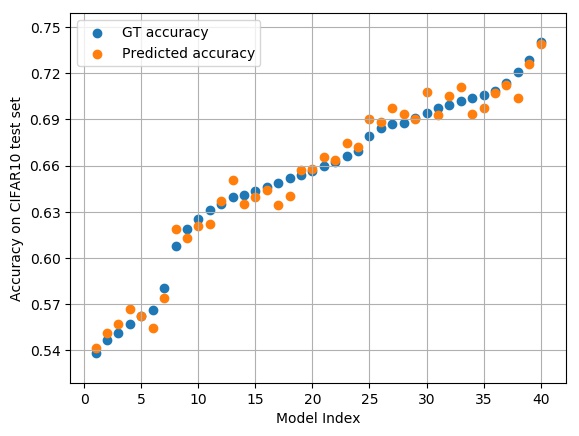}
         \caption{Using Linear Regression (D=0.5), using scheme and 9 epoch features as input, MAE=0.006.}
     \end{subfigure}
     \begin{subfigure}[b]{0.4\textwidth}
         \centering
         \includegraphics[width=\textwidth]{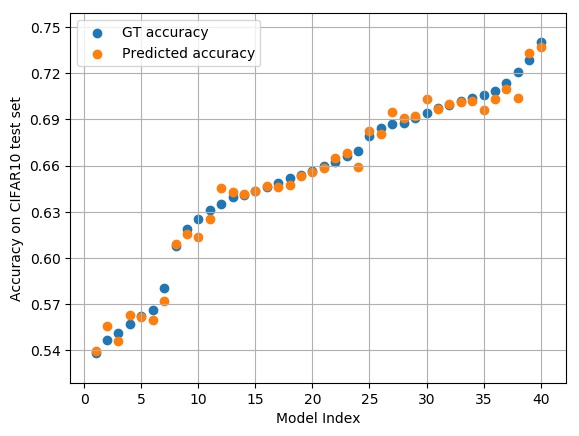}
         \caption{Using Random Forest (N=200 trees), using scheme and 9 epoch features as input, MAE=0.004.}
     \end{subfigure}
        \caption{Accuracy prediction results on the NAAP-440 test set.}
        \label{fig:predictionResults}
\end{figure}

\begin{table}[]
\centering
\resizebox{\linewidth}{!}{%
\begin{tabular}{l|cc}

\textbf{Feature Set} & \textbf{MAE} & \textbf{\#Violations} \\
\hline
All but NumMACs & 0.0071 & 50 \\
All but Depth & 0.0078 & 57 \\
All but LastLayerWidth & 0.0077 & 57 \\
All but FirstLayerWidth & 0.008 & 58 \\
All & 0.0079 & 60 \\
All but NumStages & 0.0134 & 81 \\
All but LogNumParams & 0.0199 & 111 \\
\hline
LogNumParams \& NumStages & \textbf{0.0063} & \textbf{48}\\
 
\end{tabular}
}
\caption[]{Ablation study on the scheme features using Random Forest regressor with N=200 trees. When only using scheme features (without any features from the training process), the LogNumParams and NumStages features are the most essential ones, while others do not contribute to, or even harm the performance.}
\label{tbl:schemeAblation}
\end{table}

\begin{table}[]
\centering
\resizebox{\linewidth}{!}{%
\begin{tabular}{l|ccc}

\textbf{Training} & \textbf{0 Scheme} & \textbf{2 Scheme} & \textbf{6 Scheme} \\
\textbf{Length} & \textbf{Features} & \textbf{Features} & \textbf{Features} \\
\hline
0 epochs & - & \textbf{48} & 60 \\
3 epochs & 110 & 35 & \textbf{32} \\
6 epochs & 95 & 30 & \textbf{25} \\
9 epochs & 78 & 37 & \textbf{25} \\
12 epochs & 72 & 41 & \textbf{27} \\
15 epochs & 56 & 42 & \textbf{29} \\
18 epochs & 63 & 51 & \textbf{44} \\

\end{tabular}
}
\caption[]{Ablation study on the scheme features when combining features from the initial epochs of the architectures' training processes on CIFAR10, using Random Forest regressor with N=200 trees. The values reported on the table are the number of monotonicity violations made by the regression model. The subset of 2 scheme features is only preferable when using no features from the training process.}
\label{tbl:schemeAblationWithEpochs}
\end{table}

\begin{figure*}
     \centering
     \begin{subfigure}[b]{0.19\textwidth}
         \centering
         \includegraphics[width=\textwidth]{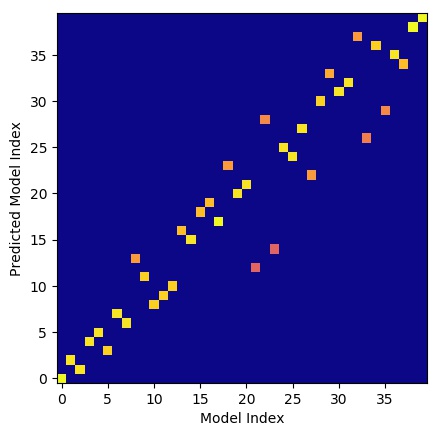}
         \caption{64 violations}
     \end{subfigure}
     \hfill
     \begin{subfigure}[b]{0.19\textwidth}
         \centering
         \includegraphics[width=\textwidth]{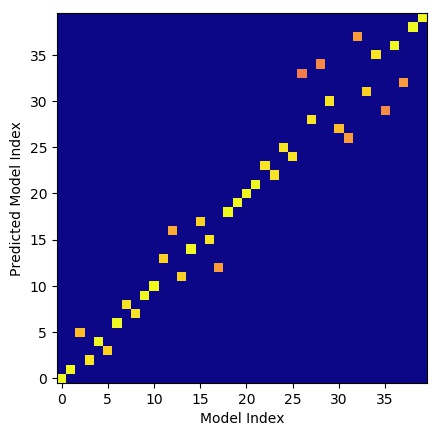}
         \caption{48 violations}
     \end{subfigure}
     \hfill
     \begin{subfigure}[b]{0.19\textwidth}
         \centering
         \includegraphics[width=\textwidth]{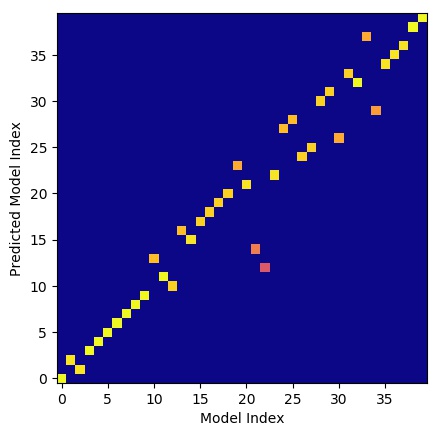}
         \caption{42 violations}
     \end{subfigure}
     \hfill
     \begin{subfigure}[b]{0.19\textwidth}
         \centering
         \includegraphics[width=\textwidth]{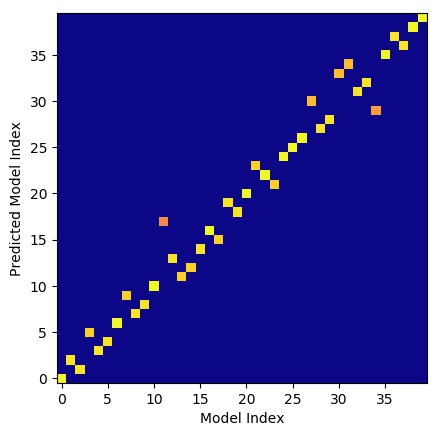}
         \caption{30 violations}
     \end{subfigure}
     \hfill
     \begin{subfigure}[b]{0.19\textwidth}
         \centering
         \includegraphics[width=\textwidth]{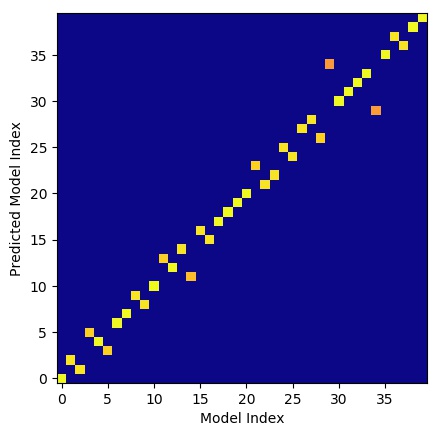}
         \caption{25 violations}
     \end{subfigure}
        \caption{Visualized monotonicity of accuracies predicted by: (a) linear regression only using scheme features, (b) linear regression using scheme features and features from 9 training epochs, (c) Gradient Boosting (N=100 estimators) only using scheme features, (d) SVR (RBF kernel) using scheme features and features from 6 training epochs, (e) Random Forest (N=200 estimators) using scheme features and features from 9 training epochs.}
        \label{fig:monotonicityResults}
\end{figure*}

\begin{table*}[]
\centering
\resizebox{\linewidth}{!}{%
\begin{tabular}{l||c|c|c|c}
\hline
\hline
 &  \multicolumn{4}{c}{\textbf{MAE / Monotonicity Score / \#Monotonicity Violations}} \\
 \hline
 & 100.0\% acceleration & 96.7\% acceleration & 93.3\% acceleration & 90.0\% acceleration\\ 
\textbf{Algorithm} & (0 epochs) & (3 epochs) & (6 epochs) & (9 epochs)\\ 
\hline
1-NN & 0.007 / 0.933 / 52  & 0.009 / 0.929 / 55  & 0.007 / 0.940 / 47  & 0.006 / 0.959 / 32 \\ 
3-NN & 0.009 / 0.918 / 64  & 0.007 / 0.944 / 44  & 0.007 / 0.950 / 39  & 0.007 / 0.951 / 38 \\ 
5-NN & 0.010 / 0.908 / 72  & 0.008 / 0.942 / 45  & 0.007 / 0.941 / 46  & 0.007 / 0.949 / 40 \\ 
7-NN & 0.009 / 0.909 / 71  & 0.007 / 0.950 / 39  & 0.007 / 0.951 / 38  & 0.006 / 0.962 / 30 \\ 
9-NN & 0.010 / 0.914 / 67  & 0.009 / 0.942 / 45  & 0.007 / 0.960 / 31  & 0.007 / 0.951 / 38 \\ 
Linear Regression & 0.017 / 0.918 / 64  & 0.009 / 0.926 / 58  & 0.008 / 0.941 / 46  & 0.007 / 0.942 / 45 \\ 
Linear Regression (D=0.5) & 0.015 / 0.919 / 63  & 0.008 / 0.932 / 53  & 0.007 / 0.942 / 45  & 0.006 / 0.947 / 41 \\ 
Linear Regression (D=0.25) & 0.013 / 0.919 / 63  & 0.008 / 0.935 / 51  & 0.007 / 0.942 / 45  & 0.006 / 0.947 / 41 \\ 
Decision Tree & 0.007 / 0.931 / 54  & 0.007 / 0.929 / 55  & 0.008 / 0.924 / 59  & 0.007 / 0.933 / 52 \\ 
Gradient Boosting (N=25) & 0.009 / 0.944 / 44  & 0.008 / 0.953 / 37  & 0.006 / 0.951 / 38  & 0.006 / 0.958 / 33 \\ 
Gradient Boosting (N=50) & 0.007 / 0.940 / 47  & 0.006 / 0.955 / 35  & 0.006 / 0.953 / 37  & 0.005 / 0.956 / 34 \\ 
Gradient Boosting (N=100) & 0.006 / \textbf{0.946 / 42}  & 0.006 / 0.958 / 33  & 0.006 / 0.960 / 31  & 0.006 / 0.959 / 32 \\ 
Gradient Boosting (N=200) & \textbf{0.005} / 0.945 / 43  & 0.006 / 0.951 / 38  & 0.006 / 0.962 / 30  & 0.006 / 0.958 / 33 \\ 
AdaBoost (N=25) & 0.010 / 0.933 / 52  & 0.009 / 0.947 / 41  & 0.007 / 0.953 / 37  & 0.006 / 0.955 / 35 \\ 
AdaBoost (N=50) & 0.010 / 0.933 / 52  & 0.008 / 0.945 / 43  & 0.006 / 0.953 / 37  & 0.006 / 0.958 / 33 \\ 
AdaBoost (N=100) & 0.010 / 0.933 / 52  & 0.008 / 0.944 / 44  & 0.007 / 0.951 / 38  & 0.005 / 0.955 / 35 \\ 
AdaBoost (N=200) & 0.010 / 0.933 / 52  & 0.008 / 0.944 / 44  & 0.007 / 0.951 / 38  & 0.006 / 0.954 / 36 \\ 
SVR (RBF kernel) & 0.009 / 0.913 / 68  & 0.007 / 0.949 / 40  & 0.005 / 0.962 / 30  & 0.005 / 0.960 / 31 \\ 
SVR (Polynomial kernel) & 0.020 / 0.911 / 69  & 0.008 / 0.940 / 47  & 0.009 / 0.919 / 63  & 0.010 / 0.918 / 64 \\ 
SVR (Linear kernel) & 0.017 / 0.917 / 65  & 0.009 / 0.933 / 52  & 0.008 / 0.944 / 44  & 0.007 / 0.947 / 41 \\ 
Random Forest (N=25) & 0.007 / 0.935 / 51  & 0.006 / 0.954 / 36  & 0.005 / 0.958 / 33  & 0.004 / 0.964 / 28 \\ 
Random Forest (N=50) & 0.006 / 0.936 / 50  & 0.006 / 0.956 / 34  & 0.005 / 0.963 / 29  & 0.005 / 0.964 / 28 \\ 
Random Forest (N=100) & 0.006 / 0.939 / 48  & \textbf{0.005} / 0.956 / 34  & 0.005 / 0.967 / 26  & 0.004 / 0.965 / 27 \\ 
Random Forest (N=200) & 0.006 / 0.939 / 48  & 0.005 / \textbf{0.959 / 32}  & \textbf{0.005 / 0.968 / 25}  & \textbf{0.004 / 0.968 / 25} \\ 

\hline
\end{tabular}
}
\caption[]{Baseline results using various regression algorithms. For linear regression with a specified degree D, the model is trained to predict $y = (\vec{w} \cdot \vec{x}+1)^D$ rather than $y = \vec{w} \cdot \vec{x}$. It is achieved by training the model to predict $y^{1/D} - 1$. All ensemble regressors are specified with their number of estimators, N.}
\label{tbl:baseline}
\end{table*}

\section{Evaluation Baseline}
\label{sec:baseline}

\paragraph{Scheme Features Ablation Study.}
As described in Section~\ref{sec:dataset}, the dataset's schematic details include 6 features, which are the network's depth, number of stages, first and last layer widths, number of total parameters and number of total MACs. We aim at checking whether all of the scheme features are essential to training a regression model. We were surprised to witness most of the scheme features actually harming the regression accuracy repeatedly, under different regression algorithms, including Random Forest, Gradient Boosting and Linear Regression. Table~\ref{tbl:schemeAblation} demonstrates that that number of parameters and number of stages are most essential for an accurate regression, as removing them from the feature set causes a notable degradation. On the other hand, the removal of each of the other features actually contributes to the regressor performance, when using a Random Forest regressor. However, when we do combine the features from the training process, which as described in Section~\ref{sec:dataset}, include each epoch's accuracy on CIFAR10 test set as well as its mean and median loss on the training set, we observe that the complete set of scheme features is essential. Table~\ref{tbl:schemeAblationWithEpochs} shows the essence of the complete set of scheme features when features from the training process are included. Furthermore, when using a linear regression model, we find it very essential to use a logarithmic version of the number-of-parameters feature. When only using scheme features, it reduces the trained regression model's number of monotonicity violations from 89 too 64 and the MAE from 2.3\% to 1.7\%. To conclude, we recommend to apply the logarithmic function on the number-of-parameters feature, to only use the limited set of scheme features when choosing not to train the model at all and to keep using all 6 scheme features when including features from the training process.

\paragraph{Baseline.}
After we have understood the effect and contribution of each feature to the performance of the regressor, we proceed to providing a baseline using many regression algorithms under different settings in Table~\ref{tbl:baseline}. We used the scikit-learn~\cite{scikit-learn} Python package for that purpose. Since some of the regression algorithms have different performances under different random seeds, we repeat running each algorithm 5 times using different seeds and report their median performance, according to the number of violations. As guided by the ablation studies, we use the limited scheme feature set when not using features from the training epochs, while using all the scheme features when we do use features from the training epochs. Furthermore, we exclude results when using features of more than 9 training epochs, since as demonstrated in Table~\ref{tbl:schemeAblationWithEpochs}, we observe that features from further training epochs usually do not contribute to and even harm the performance. Figures~\ref{fig:predictionResults},\ref{fig:monotonicityResults} demonstrate the strengths of different regression algorithms with different feature sets. In all the experiments conducted, all features are fed to the regressors after scale and bias normalization, which is learned from the training set.

\paragraph{Naive Reference.}
As a reference for the regression algorithms, we report that the MAE achieved by a dummy regressor, which always outputs the mean of the training set, is 0.043. As all architectures are predicted with the same accuracy, their prediction-based order is equivalent to any random permutation. It should be possible to prove that the average number of violations over all permutations is identically $\frac{1}{2}\binom{N}{2}$. We check 500 random permutations of the 40 architectures, and as we expect, we observe that their average number of monotonicity violations is $390 = \frac{1}{2}\binom{40}{2}$, which results in a monotonicity score of exactly 0.5. Therefore, a reasonable regressor should achieve MAE $< 0.043$ and less than 390 monotonicity violations on the NAAP-440 test set. The naive reference emphasizes the importance of using the monotonicity score alongside the MAE metric.

\section{Conclusions and Future Work}
We have presented the NAAP-440 dataset for architecture accuracy prediction, which includes 440 scheme-generated neural architectures trained and evaluated on the CIFAR10 dataset. Our experiments show that by accelerating the training process by up to 100\%, it is possible to predict an architecture's accuracy score quite precisely, as well as maintain the 40 test architecture's accuracy-based order with a quite a low number of violations. As we have published the dataset and the code, we hope that the dataset will be used by other researchers to improve and push the boundaries of neural architecture accuracy prediction capabilities.

For future work, we may consider (1) examining additional train/test divisions that may require the usage of regression algorithms that can extrapolate, (2) enriching the dataset with more architectures, (3) creating another dataset that will have a fixed architecture and variable training recipes, as opposed to this work's fixed training recipe and variable architectures.

{\small
\bibliographystyle{ieee_fullname}
\bibliography{egbib}
}

\end{document}